\def\paperID{XXXX}
\def\confName{XXXX}
\def\confYear{XXXX}

\newif\ifreview 
\newif\ifarxiv \newcommand{\arxiv}{\arxivtrue}
\newif\ifcamera 
\newif\ifrebuttal

\arxiv

\pdfoutput=1
\documentclass[10pt,twocolumn,letterpaper]{article}
\ifreview \usepackage[review]{cvpr} \fi
\ifarxiv \usepackage[pagenumbers]{cvpr} \fi
\ifrebuttal \usepackage[rebuttal]{cvpr} \fi
\ifcamera \usepackage{cvpr} \fi

\usepackage{graphicx}	
\usepackage{amsmath}	
\usepackage{amssymb}	
\usepackage{booktabs}
\usepackage{times}
\usepackage{microtype}
\usepackage{epsfig}
\usepackage{caption}
\usepackage{float}
\usepackage{placeins}
\usepackage{color, colortbl}
\usepackage{stfloats}
\usepackage{enumitem}
\usepackage{tabularx}
\usepackage{xstring}
\usepackage{multirow}
\usepackage{xspace}
\usepackage{url}
\usepackage{subcaption}
\usepackage{xcolor}
\usepackage[hang,flushmargin]{footmisc}

\usepackage[accsupp]{axessibility}

% ...

\ifarxiv  \fi

\newcommand{\R}[1]{{%
    \textbf{%
        \ifstrequal{#1}{1}{\textcolor{red}{R#1}}{%
        \ifstrequal{#1}{2}{\textcolor{blue}{R#1}}{%
        \ifstrequal{#1}{3}{\textcolor{magenta}{R#1}}{%
        \ifstrequal{#1}{4}{\textcolor{teal}{R#1}}{%
                           \textcolor{cyan}{R#1}%
        }}}}%
    }%
}}

\usepackage{xr-hyper}

\makeatletter
\newcommand*{\addFileDependency}[1]{
  \typeout{(#1)}
  \@addtofilelist{#1}
  \IfFileExists{#1}{}{\typeout{No file #1.}}
}

\makeatother

\definecolor{cvprblue}{rgb}{0.21,0.49,0.74}
\usepackage[pagebackref,breaklinks,colorlinks,allcolors=cvprblue]{hyperref}
\usepackage[capitalize]{cleveref}
\crefname{section}{Sec.}{Secs.}
\crefname{table}{Table}{Tables}
\crefname{figure}{Fig.}{Figs.}

\ifarxiv \crefname{appendix}{App.}{Apps.}
\else \crefname{appendix}{Suppl.}{Suppls.} \fi

\usepackage{graphicx}
\usepackage{amsmath, amssymb}
\usepackage[linesnumbered,algoruled,boxed,lined]{algorithm2e}
\makeatletter 
\g@addto@macro{\@algocf@init}{\SetKwInOut{Parameter}{Parameters}} 
\makeatother
\usepackage{algpseudocode}

\definecolor{cvprblue}{rgb}{0.21,0.49,0.74}
\usepackage[pagebackref,breaklinks,colorlinks,allcolors=cvprblue]{hyperref}
\usepackage{multirow}
\usepackage{amssymb}
\usepackage{pifont}
\usepackage{float}

\newcommand{\cmark}{\ding{51}}%
\newcommand{\xmark}{\ding{55}}%
\title{Hydra-NeXt: Robust Closed-Loop Driving with Open-Loop Training}
\author{
Zhenxin Li\textsuperscript{1,2*}\quad
Shihao Wang\textsuperscript{3*}\quad
Shiyi Lan\textsuperscript{4}\quad
Zhiding Yu\textsuperscript{4}\quad
Zuxuan Wu\textsuperscript{1,2}\quad 
Jose M. Alvarez\textsuperscript{4} \\
\textsuperscript{1}Institute of Trustworthy Embodied AI, Fudan University\\ 
\textsuperscript{2}Shanghai Collaborative Innovation Center of Intelligent Visual Computing\\ \textsuperscript{3}The Hong Kong Polytechnic University\quad 
\textsuperscript{4}NVIDIA \\
}

\begin{document}
\newcommand{\hydra}{Hydra-NeXt }

\maketitle
\begin{abstract}
End-to-end autonomous driving research currently faces a critical challenge in bridging the gap between open-loop training and closed-loop deployment. 
Current approaches are trained to predict trajectories in an open-loop environment, which struggle with quick reactions to other agents in closed-loop environments and risk generating 
kinematically infeasible plans due to the gap between open-loop training and closed-loop driving.
In this paper, we introduce Hydra-NeXt, a novel multi-branch planning framework that unifies trajectory prediction, control prediction, and a trajectory refinement network in one model.
Unlike current open-loop trajectory prediction models that only handle general-case planning, Hydra-NeXt further utilizes a control decoder to focus on short-term actions, which enables faster responses to dynamic situations and reactive agents. 
Moreover, we propose the Trajectory Refinement module to augment and refine the planning decisions by effectively adhering to kinematic constraints in closed-loop environments. 
This unified approach bridges the gap between open-loop training and closed-loop driving, demonstrating superior performance of 65.89 Driving Score (DS) and 48.20\% Success Rate (SR) on the Bench2Drive dataset without relying on external experts for data collection. \hydra surpasses the previous state-of-the-art by 22.98 DS and 17.49 SR, marking a significant advancement in autonomous driving.
Code will be available at \url{https://github.com/woxihuanjiangguo/Hydra-NeXt}.
\newcommand\blfootnote[1]{%
  \begingroup
  \renewcommand\thefootnote{}\footnote{#1}%
  \addtocounter{footnote}{-1}%
  \endgroup
}
\blfootnote{$^{*}$Work done during internship at NVIDIA.}

\end{abstract}    
\vspace{-0.2in}
\section{Introduction}

End-to-end autonomous driving (E2E AD)~\cite{hu2022st, wu2022trajectory, chitta2022transfuser, hu2023planning, jiang2023vad, jaeger2023hidden, wang2023drivemlm, chen2024vadv2, renz2024carllava, weng2024drive, li2024hydra} has emerged as a trending alternative to traditional perception-planning pipelines. Current research in E2E AD is divided into open-loop and closed-loop environment-oriented approaches. 

An \textbf{open-loop environment}~\cite{caesar2020nuscenes, caesar2021nuplan, Dauner2024NEURIPS, hu2023planning} refers to an autonomous system's performance being assessed without environmental feedback. The main advantage of this approach is its simplicity and efficiency, as it avoids the complexity of real-time feedback, making it cost-effective and suitable for large-scale training and testing. Benchmarks like NAVSIM~\cite{Dauner2024NEURIPS} provide controlled environments where agents follow predefined trajectories, allowing for precise evaluation of metrics like collision rates, progress, and traffic-rule following. 
However, the lack of reactive behaviors and dynamic interactions with other agents or changing conditions makes it difficult to deploy models trained in open-loop environments in real-world driving.

\begin{figure}
    \centering
        \includegraphics[width=\linewidth]{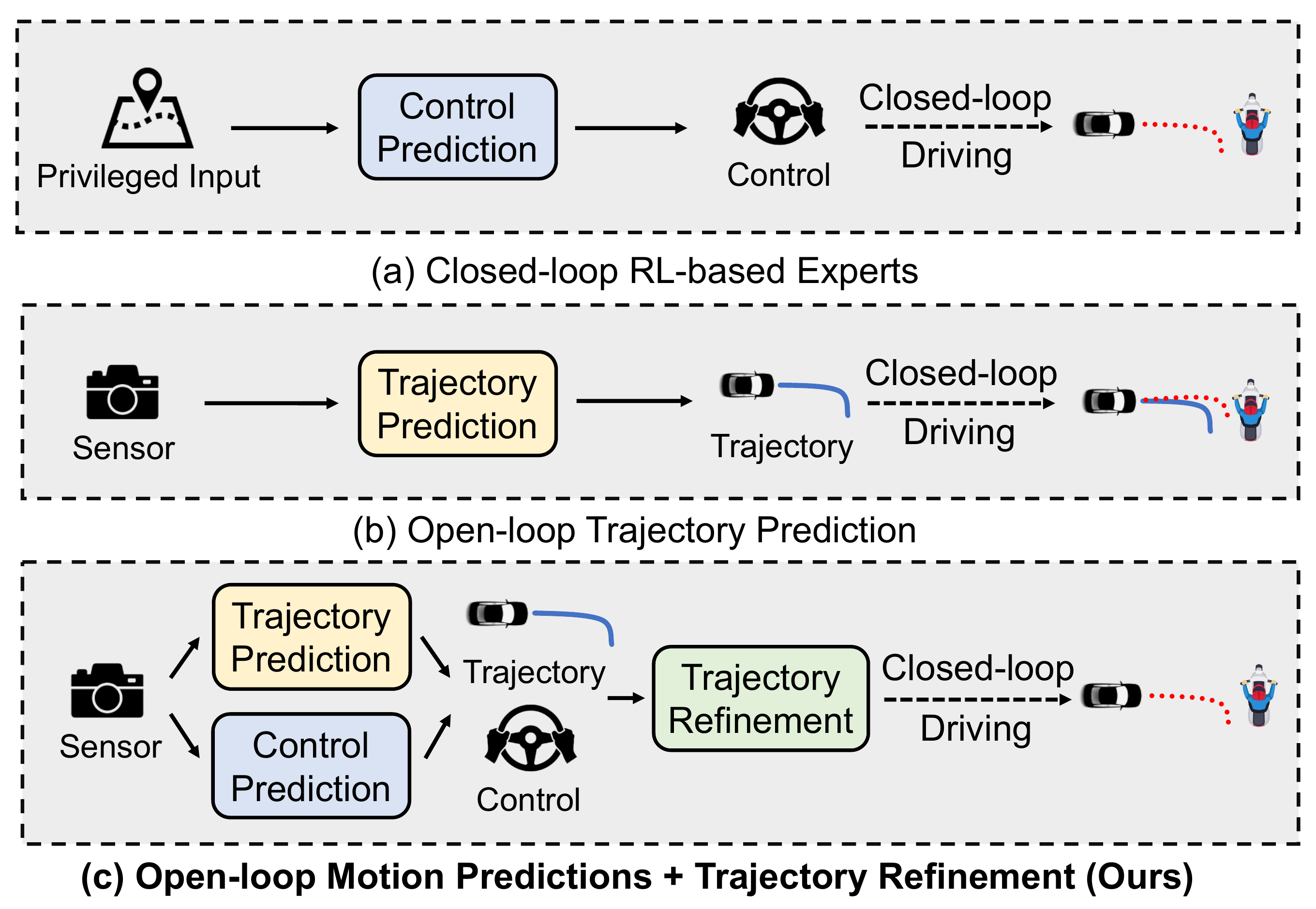}
    \vspace{-0.2in}

    \caption{\textbf{Different Paradigms of Autonomous Driving.} 
    Closed-loop RL-based experts learn from environment interactions, ensuring responsiveness to agents and kinematic feasibility by predicting control signals. End-to-end methods use open-loop trajectory prediction, often neglecting ego-agent interaction and kinematic constraints. Our approach integrates trajectory prediction, control prediction, and a trajectory refinement module to bridge the gap between open-loop training and closed-loop driving.}
    \vspace{-0.2in}
\label{fig:teaser_v2}
\end{figure}

A \textbf{closed-loop environment} in autonomous driving involves real-time feedback between the agent’s actions and the environment, typically via a simulator~\cite{CARLAv2,dosovitskiy2017carla,caesar2021nuplan}. The primary advantage of closed-loop methods is their ability to simulate dynamic, interactive environments, allowing for more realistic testing of driving policies. However, they face significant challenges: These systems often rely on synthetic simulator data, leading to a domain gap between training and real-world deployment. For instance, the previous state-of-the-art E2E AD method~\cite{jia2023driveadapter} depends heavily on reinforcement learning (RL) teacher feature embeddings, unavailable in real-world data. In addition, the driving behavior learned may deviate from human patterns~\cite{gao2024cardreamer, li2024think}, and evaluation metrics focus mainly on collision rates, overlooking other important factors such as smoothness and driving comfort~\cite{jia2024bench}.

The gap between open-loop and closed-loop environment has led to different approaches in each domain, and this separation continues to grow. 
Fig.~\ref{fig:teaser_v2} shows that closed-loop methods learn to control through reinforcement learning in simulations~\cite{jia2023driveadapter, li2024think, zhang2021end, jia2023think}, while open-loop methods focus on imitating expert trajectories~\cite{hu2023planning, jiang2023vad, chen2024vadv2}.
The gap between the two domains stem from their training methods and output representations, which can be potentially addressed through two approaches. 
We could either adapt models trained in an open-loop environment to a closed-loop environment or bring closed-loop models to open environments. 
We argue that the first option, \textit{From Open-loop to Closed-loop}, is more promising due to the availability of massive real-world data.
In contrast, the second option, \textit{From Closed-loop to Open-loop}, faces sim-to-real domain gaps~\cite{kadian2020sim2real, hofer2021sim2real, hu2023simulation}, which pose the biggest challenge to deploy a model in a real car. Hence, we focus on addressing the limitations of open-loop models when driving in closed-loop environments, which involve \textbf{handling reactive agents} and \textbf{adhering to kinematic constraints}.

As the state-of-the-art open-loop E2E method on the NAVSIM benchmark~\cite{Dauner2024NEURIPS}, Hydra-MDP shows promising results for non-reactive agents, so the challenge is to extend this ability to reactive agents. 
Non-reactive agents maintain their behavior even if the ego vehicle deviates from the ground truth trajectory. 
In contrast, reactive agents adjust in response to the ego’s predicted trajectory, potentially causing the E2E AD model's collision predictions to fail. The general issue caused by open-loop training is that most open-loop metrics~\cite{hu2023planning, caesar2021nuplan, Dauner2024NEURIPS} focus on  waypoints rather than control signals. 
While waypoints are useful for general navigation, they do not account for immediate changes in the behavior of other agents, leading to delays in reaction time when unexpected events occur. 
In contrast, control signals directly influence the vehicle's actions, such as braking or steering, allowing for faster adjustments in response to sudden changes.
Although existing literature \cite{wu2022trajectory} also predicts control with waypoints directly, it performs poorly in  interactive situations~\cite{jia2024bench}, which may stem from a lack of multi-modal decisions and insufficient exploration of their nuanced relationships.
Further, the predicted waypoints may not satisfy kinematic constraints in the closed-loop environment, which can cause compounding errors~\cite{dauner2023parting} and result in unsafe behaviors in dynamic scenarios.

Therefore, we introduce \hydra, featuring a multi-branch planning framework.
\hydra consists of Multi-head Motion Decoders for trajectory and control prediction, and a Trajectory Refinement network for kinematics-based proposal selection. 
In addition to the trajectory decoder to handle general-case planning~\cite{li2024hydra}, the control decoder enhances the capability for short-term actions. 
Additionally, the Trajectory Refinement module applies kinematic constraints to improve predictions from both decoders, effectively combining and refining the planning decisions. 

Our contributions can be summarized as follow:

\begin{enumerate}
    \item We propose \hydra, which is the first framework to unify control, trajectory prediction, and trajectory refinement. \hydra demonstrates robust closed-loop driving performance with only open-loop learning.
    \item We benchmark Hydra-NeXt against other E2E AD solutions on the Bench2Drive dataset~\cite{jia2024bench} based on the simulator CARLA v2~\cite{CARLAv2}. \hydra outperforms previous state-of-the-art methods by a significant margin (+22.9 Driving Score, +17.5 Success Rate) by the CARLA v2 evaluation protocol without relying on external experts for data collection. 
    \hydra also generalizes well to real-world data, achieving a new state-of-the-art on the open-loop planning benchmark NAVSIM~\cite{Dauner2024NEURIPS}.
\end{enumerate}

\section{Related Work}
\subsection{Open-loop End-to-end Autonomous Driving}
To explore end-to-end planning with real-world vehicle data~\cite{caesar2020nuscenes}, researchers have integrated modularized neural networks into end-to-end fully-differentiable stacks~\cite{hu2022st, hu2023planning, weng2024drive, jiang2023vad}.
This approach quickly gained significant interest in the AV community as it shows a promising direction to scalable autonomy. 
However, recent works~\cite{Dauner2024NEURIPS, dauner2023parting, li2024ego} discover heavy biases of current E2E planning datasets~\cite{Dauner2024NEURIPS, li2024ego}, the imitation learning paradigm~\cite{dauner2023parting, li2024ego}, and evaluation protocols~\cite{Dauner2024NEURIPS, li2024ego}, rendering open-loop E2E methods as unreliable. 
To address these issues, a more advanced benchmark NAVSIM~\cite{Dauner2024NEURIPS} is proposed to filter homogeneous planning data and benchmark with a series of rule-based metrics for collision avoidance, map adherence, and more.
The state-of-the-art open-loop planner Hydra-MDP~\cite{li2024hydra}, which learns these rule-based metrics via knowledge distillation, achieves greater reliability than imitation-based E2E planners.
Building on Hydra-MDP, we develop a model suited for dynamic scenarios, transitioning from open-loop trajectory planning to a closed-loop framework using control prediction and kinematics-based trajectory refinement.

\subsection{Closed-loop Driving Benchmarks and Policies}
Compared with open-loop E2E autonomous driving, closed-loop driving has a longer history.
Benchmarks such as nuPlan~\cite{caesar2021nuplan}, CARLA v1~\cite{dosovitskiy2017carla}, CARLA v2~\cite{CARLAv2}, and Bench2Drive~\cite{jia2024bench} focus on different aspects of driving policies, from trajectory planning~\cite{caesar2021nuplan} to E2E driving~\cite{dosovitskiy2017carla, CARLAv2, jia2024bench}.
Recently, CARLA v2~\cite{CARLAv2} poses significant challenges as it requires the handling of interactive and dynamic scenarios. 
Bench2Drive~\cite{jia2024bench} optimizes the variances produced by the CARLA v2 evaluation protocol and collects data from an RL-based expert~\cite{li2024think} on diverse, shorter routes, making it suitable to assess policies in varied conditions.

On these benchmarks, expert-level policies include rule-based path planners~\cite{Beißwenger2024PdmLite, dauner2023parting} and RL-based policies~\cite{zhang2021end, li2024think, gao2024cardreamer}. 
These policies take privileged inputs such as ground-truth perception and map data, while E2E AD methods~\cite{wu2022trajectory, chitta2022transfuser, jaeger2023hidden, wang2023drivemlm, chen2024vadv2, renz2024carllava} only use sensor observations. 
These E2E policies, as well as previous open-loop E2E methods, fall behind expert-level policies when it comes to dynamic and interactive scenarios~\cite{CARLAv2}. 
Moreover, Bench2Drive shows that these methods rely on expert feature embeddings~\cite{jia2023driveadapter} for better performance.
In our paper, we aim to close the gap between the two worlds: open-loop planning and closed-loop driving by enhancing a top open-loop planner~\cite{li2024hydra} with control and kinematic components.

\begin{figure}
    \centering
            \includegraphics[width=\linewidth]{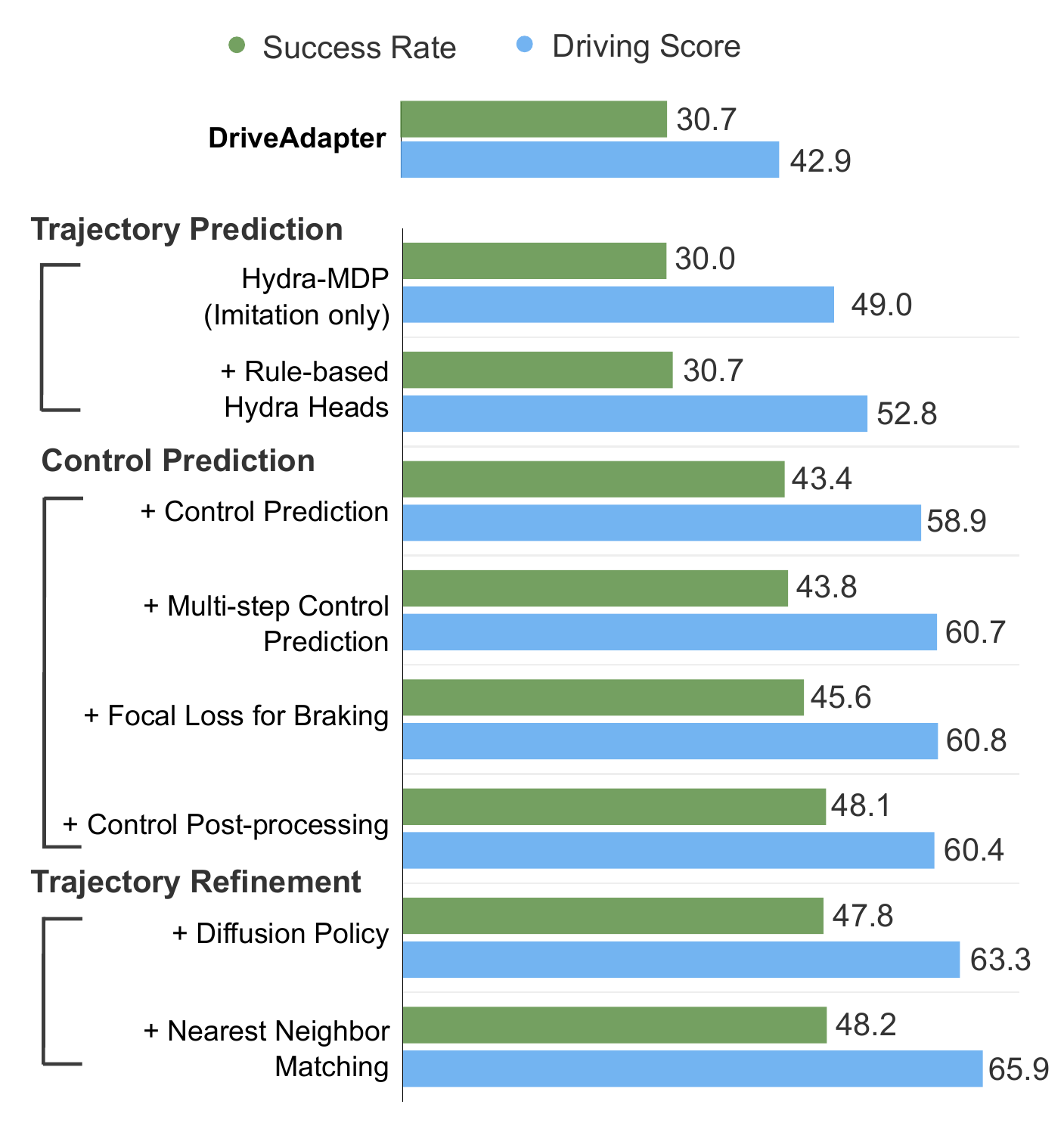}
    \vspace{-0.3in}

    \caption{
    \textbf{Roadmap from Hydra-MDP to \hydra.} DriveAdapter~\cite{jia2023driveadapter} was the previous state-of-the-art method on the Bench2Drive benchmark.
    }
    \vspace{-0.2in}
\label{fig:abl_curve}
\end{figure}

\subsection{Diffusion-based Driving Policies}
Diffusion policies~\cite{chi2023diffusion, xian2023chaineddiffuser, chen2024overview} have been used widely for generating robot behaviors.
They prove effective in capturing multi-modal action distributions and generating smooth trajectories.
In autonomous driving, recent works leverage diffusion models to predict trajectory waypoints given privileged input of surrounding traffic scenes~\cite{jiang2023motiondiffuser, guo2023scenedm, yang2024diffusion, westny2024diffusion}. 
Our approach differs by employing the diffusion policy to generate multiple smooth and high-frequency control sequences, which serve as additional proposals for trajectory refinement.

\section{Trajectory Decoder: Hydra-MDP}
\label{sec:hydramdp}
\vspace{-0.02in}
Before introducing Hydra-NeXt, we start from Hydra-MDP~\cite{li2024hydra}, a multi-modal planner with a perception network and a trajectory decoder, which learns from both human demonstrations and rule-based open-loop metrics.

\vspace{0.01in}\noindent\textbf{Perception Network:} Given front-view and back-view images, we use an image backbone to extract multi-view features. These features are flattened into a sequence of environment tokens $F_{env}$ to represent the surroundings. To focus on motion planning, we apply a minimal design in the perception network without auxiliary perception tasks~\cite{hu2023planning, weng2024drive, jiang2023vad, chen2024vadv2, chitta2022transfuser}, although they can facilitate planning.

\vspace{0.01in}\noindent\textbf{Trajectory Decoder:}
The trajectory decoder $\pi_{traj}$ generates a trajectory $T$ for routing based on environment tokens $F_{env}$ and a discrete trajectory vocabulary $\mathcal{V}$~\cite{philion2020lift, phan2020covernet, chen2024vadv2}.
Following Hydra-MDP~\cite{li2024hydra},
we create a trajectory vocabulary $\mathcal{V}$ with 4096 discrete trajectories, embed them into latent queries $Q_{traj}$, and attend to environment tokens $F_{env}$ in a transformer decoder~\cite{vaswani2017attention}.
With only imitation learning, this approach achieves 49.0 DS as shown in Fig.~\ref{fig:abl_curve}, outperforming the previous state-of-the-art DriveAdapter by 6.1 DS.
Further, we implement an open-loop metric system on Bench2Drive using the following metrics:
\begin{itemize}
  \setlength\itemsep{0.05in}
    \item \textbf{Collision:} The collision metric checks if the ego vehicle intersects with other agents in the Bird's-Eye View (BEV) space~\cite{hu2023planning, jiang2023vad, li2024ego, Dauner2024NEURIPS}. We compute collisions at a higher frequency (\ie 10Hz) by interpolating trajectory waypoints following NAVSIM~\cite{Dauner2024NEURIPS}.

    \item \textbf{Soft Lane Keeping:} To ensure lane adherence while allowing lane changing, we design a soft lane keeping metric that bounds the ego's driving area following~\cite{jiang2023vad}. We penalize trajectories with excessive angular differences between trajectory segments and lane segments.

    \item \textbf{Ego Progress:} The Ego Progress metric is included to discourage passive driving behaviors and promote driving progress. Similar to~\cite{Dauner2024NEURIPS, caesar2021nuplan}, we project the trajectory waypoints onto the lane segments that the expert travels through, and normalize the ego distance by the expert's.
\end{itemize} \leavevmode
\begin{figure*}[tp]
    \centering
    \includegraphics[width=\linewidth]{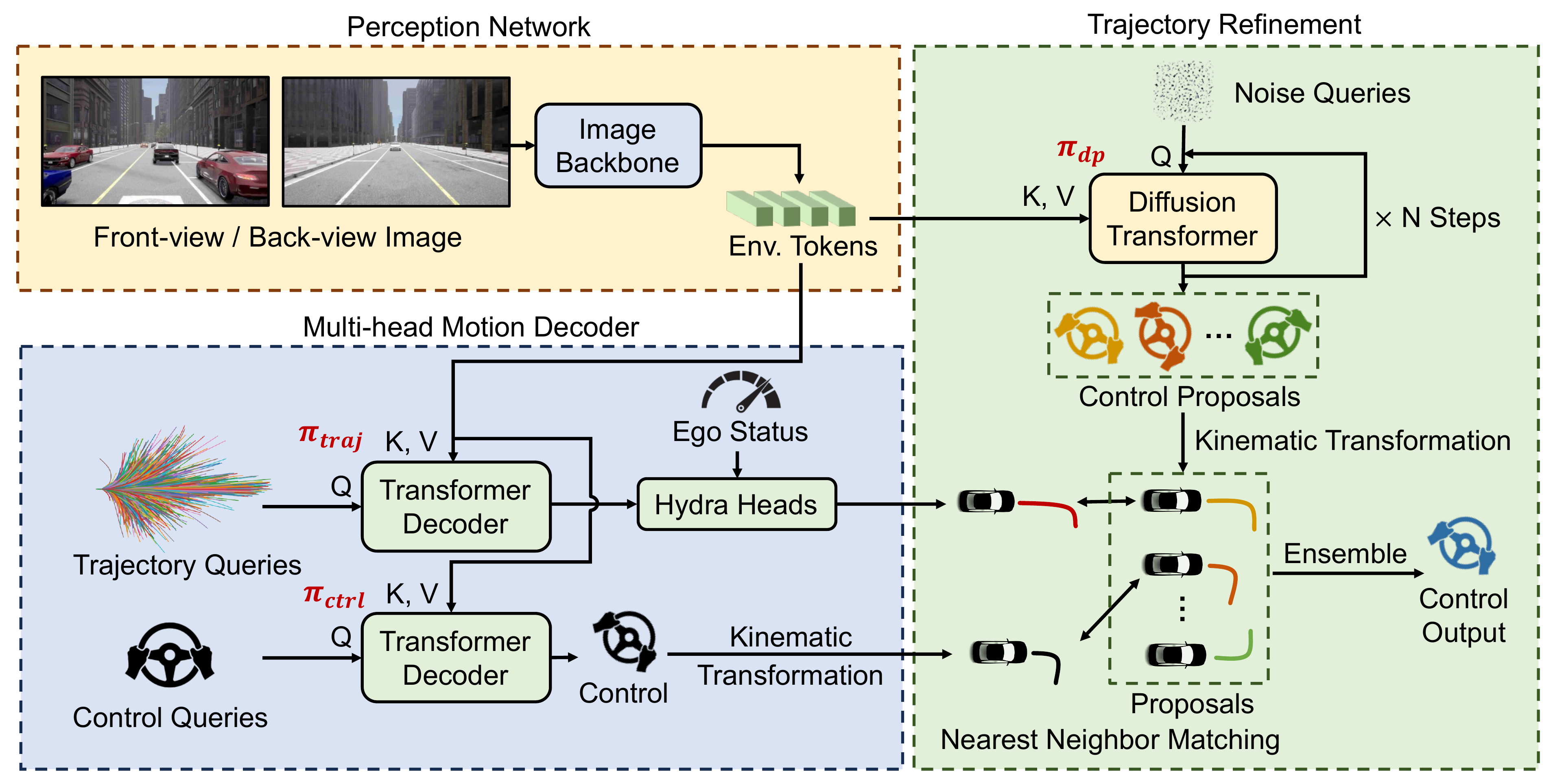}
    \vspace{-0.3in}

    \caption{\textbf{The Overall Architecture of \hydra.}
    The Perception Network first processes raw sensor observations into environment tokens.
    The Multi-head Motion Decoder (Trajectory Decoder $\pi_{traj}$ and Control Decoder $\pi_{ctrl}$) is responsible for generating a trajectory and a control tuple based on these tokens, which are refined by the Trajectory Refinement Module $\pi_{dp}$ into final control signals.
    }
    \vspace{-0.15in}

    \label{fig:arch}
\end{figure*}

\noindent These metrics are used as rule-based teachers in Hydra-MDP. Fig.~\ref{fig:abl_curve} shows the performance enhancement brought by rule-based heads, which appear to be limited compared with observations on NAVSIM~\cite{Dauner2024NEURIPS}.
This is likely caused by randomly disappearing agents in 
Bench2Drive~\footnote{https://github.com/Thinklab-SJTU/Bench2Drive/issues/31}, which confuse the rule-based heads. \eg A trajectory that extends far into the future may be labeled as safe if its future waypoints do not intersect with a disappeared agent, even though the agent is currently visible and close to the ego vehicle. 
In Sec.~\ref{sec:hydranext}, we propose Hydra-NeXt, an extended version of Hydra-MDP addressing the limitations in closed-loop driving.

\section{Hydra-NeXt}
\label{sec:hydranext}
In this section, we elaborate on Hydra-NeXt, an E2E framework with robust closed-loop abilities. 
Apart from the Trajectory Decoder $\pi_{traj}$, Hydra-NeXt has two other policies: \textbf{Control Decoder $\pi_{ctrl}$}, and \textbf{Trajectory Refinement $\pi_{dp}$}.

\subsection{Control Decoder $\pi_{ctrl}$}
Our goal is to enable quick responses to reactive agents through direct control output, instead of relying on trajectory waypoints for rapid decisions. This is because waypoints are typically transformed into control signals through specific controllers such as a PID Controller or Model Predictive Control~\cite{camacho2013model}, which can lead to errors during path following.
Further, TCP~\cite{wu2022trajectory} points out that ensembling the trajectory and the control signal can boost driving performance, but it only considers a single modality for both the trajectory and the control prediction branches. This limitation makes it challenging to address uncertainties in environments~\cite{chen2024vadv2} and susceptible to interpolate between different action modes~\cite{jaeger2023hidden}. 

Based on these findings, we add a second classification-based control decoder $\pi_{ctrl}$ to generate control signals $(C^1,...,C^{t_{ctrl}})$ for $t_{ctrl}$ timesteps. In the CALRA simulator~\cite{dosovitskiy2017carla, CARLAv2}, each control signal is a tuple of $(brake, throttle, steer)$ used to direct the vehicle. Meanwhile, we incorporate the idea of discretization into the control decoder $\pi_{ctrl}$ to handle uncertainty, following the practice in the RL-based driving policy Think2Drive~\cite{li2024think}.
The use of learning-based control prediction sidesteps the need to convert trajectories into control signals through traditional controllers, enabling faster responses to reactive agents in the closed-loop environment.

In particular, we first randomly initialize $t_{ctrl}$ control queries $Q_{ctrl}$, which correspond to the $t_{ctrl}$ current and future steps for which we want to predict control signals.
Similarly, $Q_{ctrl}$ attends to $F_{env}$ in a transformer decoder. 
After this, $Q_{ctrl}$ is processed by three separate MLP layers for each control signal (\ie $brake, throttle, steer$). 
Since the expert data collected by Think2Drive only consists of discrete control signals, the control decoder can be easily trained using cross-entropy loss functions without further discretizing the control signal ground truths. 
We select the control tuples with the highest likelihood at each step $(C^1,...C^{t_{ctrl}})$ as the final output of $\pi_{ctrl}$.
The design choices of the control decoder are discussed in Sec.~\ref{sec:ablate}, such as  architectures and loss functions.

\subsection{Trajectory Refinement $\pi_{dp}$}
\label{sec:traj_refine}
Another important aspect of closed-loop driving is to ensure the smoothness of driving.
Benchmarks like nuPlan~\cite{caesar2021nuplan} and NAVSIM~\cite{Dauner2024NEURIPS} post-process trajectories using an LQR tracker~\cite{tassa2014control} and a kinematic bicycle model~\cite{rajamani2011vehicle, polack2017kinematic} to adhere to kinematic constraints. 
Nevertheless, this idea has been neglected by existing open-loop E2E AD methods, which mainly focuses on cloning trajectory waypoints without considering kinematic constraints in closed-loop environments.
This oversight becomes particularly critical when the vehicle experiences control loss in closed-loop environments~\cite{CARLAv2, jia2024bench} or encounters large compounding errors~\cite{dauner2023parting} when quickly recovering to predicted waypoints is difficult.

To address this issue, we introduce the Trajectory Refinement network to augment and then combine the planning output $T$ and $C$ while following kinematic constraints. 
Inspired by PDM-Lite~\cite{Beißwenger2024PdmLite}, we use a kinematic bicycle model~\cite{rajamani2011vehicle, polack2017kinematic} that operates on high-frequency control signals for fine-grained waypoint rollouts~\cite{Beißwenger2024PdmLite}. Unlike predicted waypoints from $\pi_{traj}$, these waypoints are kinematically feasible. 
However, we empirically find that existing control policies struggle to produce a long and smooth control sequence at a high frequency (see  Sec.~\ref{sec:ablate}).
Therefore, we resort to the Diffusion Policy~\cite{chi2023diffusion}, which proves capable of generating smooth and diverse robot behaviors.
During training, a diffusion transformer $\pi_{dp}$ is trained to predict noise added to the ground-truth control sequence following the common practice of Denoising Diffusion Probabilistic Models (DDPMs)~\cite{ho2020denoising}.
During inference, $\pi_{dp}$ gradually denoises $N$ random noise queries $Q_{dp}$ into smooth action sequences $\{(\Tilde{C}^1_i,...,\Tilde{C}^{t_{dp}}_i)\}_{i=1}^N$ through numerous denoising iterations. Each proposal spans across $t_{dp}$ timesteps.

With these smooth control sequence proposals, we devise a process named Nearest Neighbor Matching for choosing the best proposal in a kinematically feasible way.
Algorithm~\ref{algo:traj} depicts the nearest neighbor matching process for proposal selection in detail.
Specifically, we first use a kinematic bicycle model to transform control sequences from $\pi_{dp}$ and $\pi_{ctrl}$ into trajectory waypoints.
After the transformation, we select two nearest control candidates to match current candidates based on L2 distances. 
The new control candidates can be viewed as a kinematically feasible version of the predictions given by the Multi-head Motion Decoder.
Finally, we ensemble the candidates $\mathcal{C}$ into the final control signal $C^*$ by averaging the throttle and steer values, which is a simplified ensembling method from TCP~\cite{wu2022trajectory}. The brake is set to 1 if the condition $\sum_{C\in \mathcal{C}} C.brake \geq \tau$ holds; otherwise, it is set to 0. Note that the brake values produced by the candidates are binary and $\tau$ is a predefined brake threshold.

\begin{algorithm}
    \caption{Nearest Neighbor Matching}
        \label{algo:traj}
        \KwIn {trajectory $T$, control sequence $(C^1,...,C^{t_{ctrl}})$, $N$ control proposals $\{(\Tilde{C}^1_i,...,\Tilde{C}^{t_{dp}}_i)\}_{i=1}^N$}
        \KwOut {control candidates $\mathcal{C}$}
        \SetKwInOut{Definitions}{Definitions}
        \Definitions {\texttt{KB}: Kinematic Bicycle Model \\ \texttt{PID}: PID controller \\
        \texttt{L$_2$}: L$_2$ Distance}

        \BlankLine
        $\mathcal{C} \gets \{\texttt{PID}(T),C^1\}$
        
        $T_{ctrl} \gets \texttt{KB}(C^1,...,C^{t_{ctrl}})$
        
        \For{$i \gets 1 \text{ to } N$}{
            $T_{dp}^i \gets KB(\Tilde{C}^1_i,...,\Tilde{C}^{t_{dp}}_i)$
        }

        $i \gets \underset{i}{\arg \min}\,\texttt{L$_2$}(T^i_{dp}, T)$

        $j \gets \underset{j}{\arg \min} \,\texttt{L$_2$}(T^j_{dp}, T_{ctrl})$
        
        $\mathcal{C} \gets \mathcal{C} \cup \{\Tilde{C}^1_i, \Tilde{C}^1_j$\}

\end{algorithm}

\vspace{-0.05in}
\section{Experiments}
\begin{table*}[t]

\centering
\small

\begin{tabular}{l|c|c|cc}
\toprule
\multirow{2}{*}{\textbf{Method}} & \multirow{2}{*}{\textbf{Sensors}}  & 
\textbf{Open-loop Metric}
& \multicolumn{2}{c}{\textbf{Closed-loop Metrics (CARLA v2)}}
\\  \cmidrule{4-5} & & L2 (meter) $\downarrow$
                                    & Driving Score $\uparrow$  & Success Rate (\%) $\uparrow$
                                    \\ 
\midrule
UniAD-Base~\cite{hu2023planning}    & 6 Cameras & \textbf{0.73}     & 37.72     & 9.54    \\
VAD~\cite{jiang2023vad}             & 6 Cameras   & 0.91   & 39.42     & 10.00  \\
Hydra-MDP~\cite{li2024hydra}        & Front/Back Cameras   & 0.82   & 52.80     & 30.73  \\
\midrule
TCP*~\cite{wu2022trajectory}  & Front Camera      & 1.70   & 23.63     & 7.72       \\
TCP-Ctrl*~\cite{wu2022trajectory}    & Front Camera  & -   & 18.63     & 5.45       \\
TCP-Traj*~\cite{wu2022trajectory}  & Front Camera   & 1.70 & 36.78     & 26.82      \\
ThinkTwice*~\cite{jia2023think}   & 6 Cameras     & 0.95   &  39.88    & 28.14     \\ 
DriveAdapter*~\cite{jia2023driveadapter} & 6 Cameras  &  1.01 & 42.91  & 30.71     \\ 
\midrule
\rowcolor{gray!20}
\textbf{\hydra }  & Front/Back Cameras  &    0.92    &  \textbf{65.89}     & \textbf{48.20}   \\ 
\bottomrule
\end{tabular}

\vspace{-0.1in}

\caption{
\textbf{Open-loop and Closed-loop Performance of E2E-AD Methods on the Bench2Drive Benchmark with the CARLA v2 Evaluation Protocol}. 
The CARLA v2 protocol calculates the Driving Score (DS) by aggregating all infractions multiplicatively, including minimum speed infractions.
* The model benefits from expert feature distillation. }
\vspace{-0.1in}
\label{tab:ds}
\end{table*}

\begin{table*}[t]

\centering
\small
\resizebox{0.83\linewidth}{!}{
\begin{tabular}{l|c|cccc}
\toprule
\multirow{2}{*}{\textbf{Method}} & \multirow{2}{*}{\textbf{Sensors}}  
 & \multicolumn{4}{c}{\textbf{Closed-loop Metrics (Bench2Drive)}} 
\\  \cmidrule{3-6} & & Driving Score $\uparrow$  & Success Rate (\%) $\uparrow$ & Efficiency $\uparrow$  & Comfort $\uparrow$
                                    \\ 
\midrule
UniAD-Base~\cite{hu2023planning}    & 6 Cameras    & 45.81     & 16.36  & 129.21 & 43.58  \\
VAD~\cite{jiang2023vad}             & 6 Cameras     & 42.35     & 15.00  &  157.94 & 46.01   \\
Hydra-MDP~\cite{li2024hydra} & Front/Back Cameras   & 59.95     & 29.82  &  186.83 & 18.62   \\
\midrule
TCP*~\cite{wu2022trajectory}  & Front Camera          & 40.70     & 15.00  &  54.26 & 47.80    \\
TCP-Ctrl*~\cite{wu2022trajectory}    & Front Camera      & 30.47     & 7.27   & 55.97 & \textbf{51.51}   \\
TCP-Traj*~\cite{wu2022trajectory}  & Front Camera    & 59.90     & 30.00  & 76.54 & 18.08   \\
ThinkTwice*~\cite{jia2023think}   & 6 Cameras        & 62.44     & 31.23  & 69.33 & 16.22  \\ 
DriveAdapter*~\cite{jia2023driveadapter} & 6 Cameras    & 64.22     & 33.08  & 70.22 & 16.01  \\ 
\midrule
\rowcolor{gray!20}
\textbf{\hydra }  & Front/Back Cameras    &  \textbf{73.86}     & \textbf{50.00} & \textbf{197.76} & 20.68    \\ 
\bottomrule
\end{tabular}
}

\vspace{-0.1in}

\caption{
\textbf{Closed-loop Performance of E2E-AD Methods on the Bench2Drive Benchmark with the Bench2Drive Evaluation Protocol}. 
* The model benefits from expert feature distillation. }
\vspace{-0.2in}

\label{tab:ds_new}
\end{table*}

\subsection{Dataset and Metrics}
We use the E2E driving benchmark Bench2Drive~\cite{jia2024bench} for training and evaluating Hydra-NeXt.
The training data in this paper uses the official training dataset of Bench2Drive, which includes 2 million frames encompassing 44 interactive scenarios. 
These data are collected by Think2Drive~\cite{li2024think}, an RL-based expert model on CARLA v2~\cite{CARLAv2}.
For evaluation, the E2E AD model is deployed in the CARLA simulator~\cite{dosovitskiy2017carla} to perform closed-loop driving on 220 short routes designed by Bench2Drive. These short routes assess the AD model's abilities on different scenarios, while ensuring low variance in the final score.

Bench2Drive includes several metrics, including Driving Score (DS), Success Rate (SR), Efficiency, Comfort, and Multi-ability Results.
The calculation of DS can be based on two evaluation protocols: CARLA v2 and Bench2Drive.
The former accumulates the minimum speed infractions into the DS, whereas the Bench2Drive protocol separates this into the metric Efficiency.
Meanwhile, the latter protocol relaxes the time constraint of the closed-loop evaluation, which reduces the difficulty for the model to complete the routes.
Without further notations, we default to the CARLA v2 evaluation protocol in ablation studies to reflect the comprehensive performance of the method using the DS metric.

Furthermore, we evaluate Hydra-NeXt on the real-world NAVSIM Benchmark~\cite{Dauner2024NEURIPS}, which evaluates a 4-second trajectory using open-loop metrics: No at-fault Collisions (NC), Drivable
Area Compliance (DAC), Time-to-collision (TTC), Comfort
(C), and Ego Progress (EP). The PDM score (PDMS) is an aggregate of these sub-metrics.
\begin{table*}[t]
\centering
\small
\begin{tabular}{l|ccccc|c}
\toprule
\multirow{2}{*}{\textbf{Method}} & \multicolumn{5}{c}{\textbf{Ability} (\%) $\uparrow$}                                                                                                                \\ \cmidrule{2-7} 
                                 & \multicolumn{1}{c}{Merging} & \multicolumn{1}{c}{Overtaking} & \multicolumn{1}{c}{Emergency Brake} & \multicolumn{1}{c}{Give Way} & Traffic Sign & \textbf{Mean} \\ \midrule
UniAD-Base~\cite{hu2023planning}       & 14.10        & 17.78           & 21.67       &  10.00         & 14.21    & 15.55        \\ 
VAD~\cite{jiang2023vad}                 & 8.11        & 24.44           & 18.64       &  20.00         & 19.15    & 18.07       \\ 
Hydra-MDP~\cite{li2024hydra}           & 19.23       & 20.00           & 45.00       &  50.00         & 41.58    & 35.16      \\ 

\midrule
TCP*~\cite{wu2022trajectory}           & 16.18        & 20.00           & 20.00       &  10.00        & 6.99    & 14.63        \\ 
TCP-Ctrl*~\cite{wu2022trajectory}       & 10.29       & 4.44           & 10.00       &  10.00        & 6.45    & 8.23        \\ 
TCP-Traj*~\cite{wu2022trajectory}       & 8.89       & 24.29           & 51.67       &  40.00        & 46.28    & 34.22        \\ 
ThinkTwice*~\cite{jia2023think}        & 27.38       &  18.42          & 35.82       &  \textbf{50.00}    & 54.23    & 37.17        \\ 
DriveAdapter*~\cite{jia2023driveadapter} & 28.82       &26.38          & 48.76      &  \textbf{50.00}    & \textbf{57.21}   & 42.08        \\ 
\midrule
\rowcolor{gray!20}
\textbf{\hydra}    & \textbf{40.00}    &\textbf{64.44}      & \textbf{61.67}  &  \textbf{50.00}    & 50.00   & \textbf{53.22}        \\
\bottomrule
\end{tabular}
\vspace{-0.1in}
\caption{\textbf{Multi-Ability Results of E2E-AD Methods on the Bench2Drive Benchmark.}  * denotes expert feature distillation.}
\vspace{-0.2in}
\label{tab:multi}
\end{table*}

\subsection{Implementation Details}
The implementation of \hydra is largely consistent with open-loop E2E baselines~\cite{jiang2023vad, hu2023planning} on Bench2Drive.
First, we train \hydra on the training data for 20 epochs with a total batch size of 256, using 8 NVIDIA V100 GPUs.
The AdamW~\cite{loshchilov2017decoupled} optimizer is used with the Cosine Annealing Scheduler~\cite{loshchilov2016sgdr} at a learning rate of $2\times 10^{-4}$ and a weight decay of 0.01.
During training, data augmentations such as random cropping and photometric distortion are applied to the input images, which are first resized to $800\times 450$.

\hydra employs an ImageNet-pretrained~\cite{deng2009imagenet} ResNet-50~\cite{he2016deep} as the image backbone to extract front-view and back-view image features.
Following VAD~\cite{jiang2023vad}, the trajectory decoder $\pi_{traj}$ predicts a 3-second trajectory at 2Hz. 
The frequencies of $\pi_{ctrl}$ and $\pi_{dp}$ are set to 2Hz and 10Hz by default, which will be further analyzed in Sec.~\ref{sec:ablate}. 
The ego status feature includes the current longitudinal velocity and a one-hot navigation command.
In Trajectory Refinement, the brake threshold $\tau$ is set to half of the candidates set size 2 for moderate behaviors, while the kinematic bicycle model is based on the implementation of PDM-Lite~\cite{Beißwenger2024PdmLite} and the denoising timestep of $\pi_{dp}$ is set to 100~\cite{chi2023diffusion}.
For NAVSIM, we extend Hydra-MDP~\cite{li2024hydra} by incorporating acceleration and steering rate prediction as a substitute for control signals. Details can be found in the appendix.

\subsection{Main Results}
As shown in Tab.~\ref{tab:ds} and Tab.~\ref{tab:ds_new}, \hydra surpasses all E2E methods on Bench2Drive in key metrics such as the DS and SR distinctively. 
Specifically, under the CARLA v2 protocol, \hydra outperforms the previous state-of-the-art, DriveAdapter~\cite{jia2023driveadapter}, by 22.98 DS and 17.49\% SR.
Under the Bench2Drive protocol, it achieves improvements of 9.64 DS and 16.92 SR, despite DriveAdapter utilizing expert features from Think2Drive~\cite{li2024think}. 
Additionally, \hydra demonstrates higher efficiency than the baselines, especially those using expert features. 
Nevertheless, \hydra falls behind TCP~\cite{wu2022trajectory} in terms of Comfort, which is likely due to frequent braking during interactive situations.

\begin{table}[h]

\centering
\fontsize{7pt}{8pt}
\selectfont
\setlength{\tabcolsep}{5pt}
\begin{tabular}{l|cccccc}
\toprule
\textbf{Method} & \textbf{NC $\uparrow$} & \textbf{DAC $\uparrow$} & \textbf{TTC $\uparrow$} & \textbf{EP $\uparrow$} & \textbf{C $\uparrow$} & \textbf{PDMS $\uparrow$} \\ \midrule
Transfuser~\cite{chitta2022transfuser} & 97.7& 92.8& 92.8& 79.2& \textbf{100}& 84.0 \\ 

Hydra-MDP~\cite{li2024hydra} & \textbf{98.3}& 96.0& 94.6& 78.7& \textbf{100}& 86.5 \\ 
DiffusionDrive~\cite{liao2024diffusiondrive} & 98.2& 96.2& \textbf{94.7}& \textbf{82.2}& \textbf{100}& 88.1 \\ 
\midrule
\rowcolor{gray!20}
\textbf{Hydra-NeXt}    & 98.1& \textbf{97.7}& 94.6& 81.8& \textbf{100}& \textbf{88.6} \\ 
\bottomrule

\end{tabular}
\vspace{-0.1in}
\caption{\textbf{Performance of E2E-AD Methods on NAVSIM.}}
\vspace{-0.2in}
\label{tab:navsim}
\end{table}

For the Multi-ability Results shown in Tab.~\ref{tab:multi}, \hydra shows an improvement of 11.14\% over DriveAdapter in the average performance.
It is worth noting that \hydra exhibits superior performance in interactive scenarios such as merging (+11.18\%), overtaking (+38.06\%), and emergency braking (+12.91\%), which indicates its proficiency in handling reactive agents. However, \hydra falls behind DriveAdapter in scenarios where the vehicle should adhere to traffic signs. This phenomenon may be caused by causal confusion~\cite{jia2023driveadapter} when encountering mixed expert behaviors before traffic signs, and is worth future investigations.
Notably, all methods achieve less satisfactory results in yielding to specialized vehicles (\eg ambulances), which may be caused by the scarcity of such training data.

Moreover, Tab.~\ref{tab:navsim} shows that \hydra achieves a new state-of-the-art on the NAVSIM Benchmark.
Though NAVSIM only evaluates the predicted trajectory from $\pi_{traj}$, the extension of learning targets in $\pi_{ctrl}$ and $\pi_{dp}$ can benefit the trajectory prediction, leading to an improvement of 2.1 PDMS over Hydra-MDP~\cite{li2024hydra}.

\subsection{Ablation Study}
\label{sec:ablate}
\vspace{0.05in}\noindent\textbf{Ablation on Different Components and their Designs.}
Fig.~\ref{fig:abl_curve} shows the detailed ablations of each component in Hydra-NeXt.
Operating on perspective image tokens, Hydra-MDP achieves a 52.8 DS, which already surpasses DriveAdapter by around 10 DS.
Nevertheless, the trajectory prediction remains insufficient when facing more complex agent interactions, which promotes the utilization of a control prediction network $\pi_{ctrl}$ for rapid reactions.
The output of $\pi_{ctrl}$ is ensembled with the trajectory in a similar fashion to Sec.~\ref{sec:traj_refine}, where the brake threshold is correspondingly set to 1 given two proposals.
This gives us a large increase of 12.7\% in the SR, indicating fewer collisions with reactive agents.
Furthermore, using multi-step control predictions into future frames as in~\cite{wu2022trajectory} provides auxiliary supervision (+1.8 DS). Applying a binary focal loss~\cite{lin2017focal} for brake prediction balances the training (+1.8\% SR) and simply post-processing the final control output, such as slowing down during turns, leads to further enhancements (+2.5\% SR).
Finally, we build the Trajectory Refinement network on top of the previous network. Involving the Diffusion Policy and applies a straightforward ensembling leads to an increase of around 3 DS, while applying the Nearest Neighbor Matching (Algorithm~\ref{algo:traj}) to follow kinematic constraints boosts the final performance to 65.9 DS and 48.2\% SR.

\begin{table}[t]

\centering
\small
\resizebox{\columnwidth}{!}{

\begin{tabular}{c|cc}
\toprule
\textbf{Policy in Traj. Refine.}
 & Driving Score $\uparrow$  & Success Rate (\%) $\uparrow$
                                    \\ 
\midrule

$\pi_{ctrl}'$     &   60.15  &  42.25  \\ 
$\pi_{dp}$    &  \textbf{65.89}     & \textbf{48.20}   \\ 
\bottomrule
\end{tabular}

}
\vspace{-0.1in}
\caption{
\textbf{
Ablation of the Policy in Trajectory Refinement.
}
}
\vspace{-0.2in}
\label{tab:abl_dp}
\end{table}
\begin{table}[t]

\centering
\small
\resizebox{\columnwidth}{!}{

\begin{tabular}{cc|cc}
\toprule
\textbf{$\pi_{ctrl}$} & 
\textbf{$\pi_{dp}$} & 
 Driving Score $\uparrow$  & Success Rate (\%) $\uparrow$ \\ 

\midrule
2 Hz   & 2 Hz   &    61.95   &  39.05  \\ 
2 Hz   & 10 Hz  & \textbf{65.89}     & \textbf{48.20}  \\ 
10 Hz  & 10 Hz  &  64.66   &  44.45  \\ 

\bottomrule
\end{tabular}

}
\vspace{-0.1in}
\caption{
\textbf{
Ablation of the Prediction Frequency of the Control Decoder and the Diffusion Policy.
}
}
\vspace{-0.1in}
\label{tab:abl_ctrl_dp_freq}
\end{table}
\begin{table}[t]

\centering
\small
\resizebox{\columnwidth}{!}{

\begin{tabular}{c|cc}
\toprule
\textbf{Proposal Number $N$}
 & Driving Score $\uparrow$  & Success Rate (\%) $\uparrow$
                                    \\ 
\midrule

5     &  65.52     & 46.36  \\ 
10    &  65.89     & \textbf{48.20}   \\ 
20    &   \textbf{66.09} &  46.08  \\ 
\bottomrule
\end{tabular}

}
\vspace{-0.1in}
\caption{
\textbf{
Ablation of the Proposal Number N in Trajectory Refinement.
}
}
\vspace{-0.25in}
\label{tab:abl_n_prop}
\end{table}
\begin{figure*}[tp]
    \centering
        \includegraphics[width=0.95\linewidth]{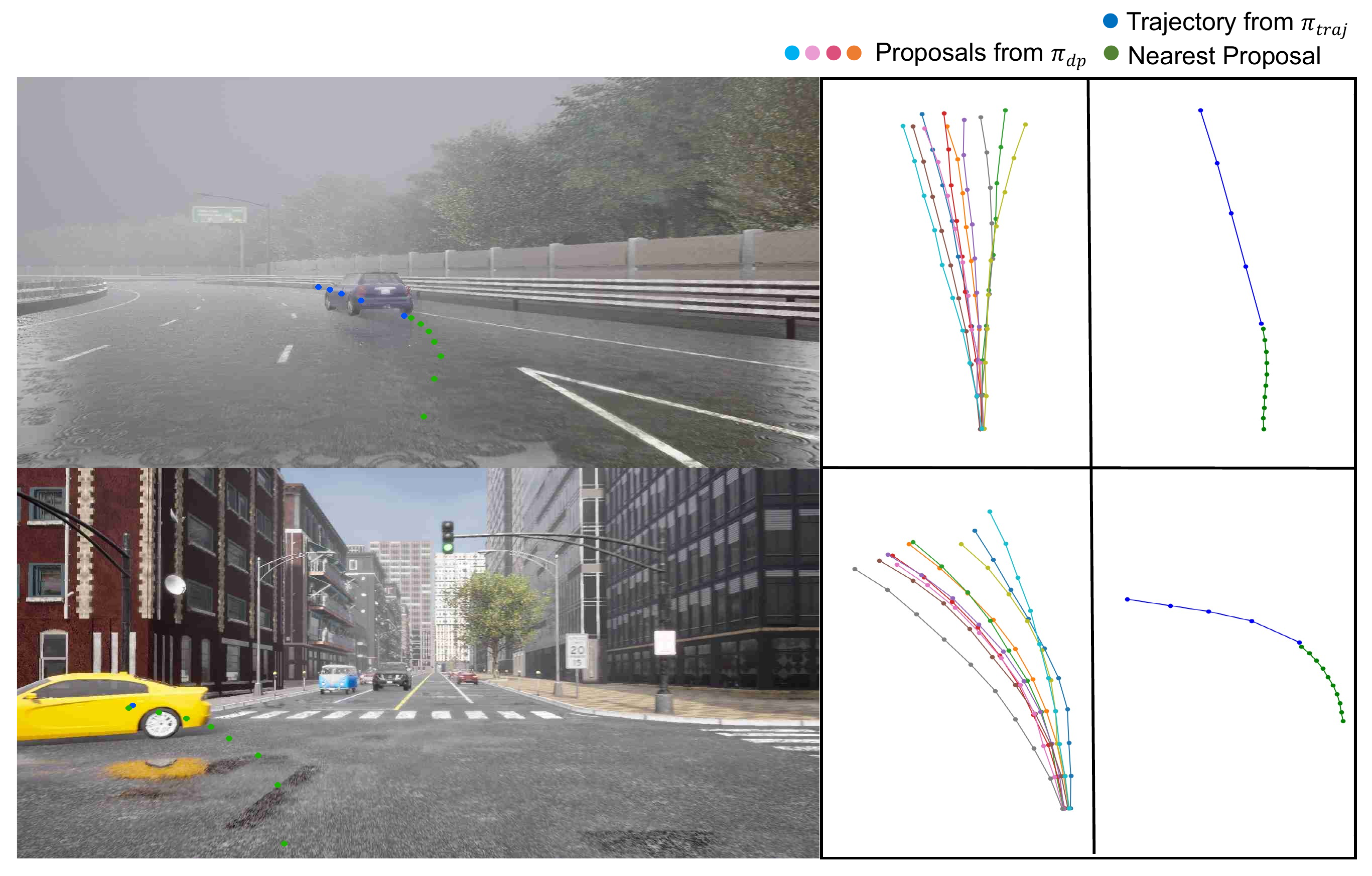}
    \vspace{-0.2in}
    \caption{\textbf{Visualizations of Trajectory Refinement.} The figure shows the front-view image, trajectories transformed from control proposals, and the selected nearest proposal to the predicted trajectory.
    }
    \vspace{-0.2in}

    \label{fig:vis}
\end{figure*}

\vspace{0.05in}\noindent\textbf{Design of Trajectory Refinement.} 
To illustrate the necessity of using the Diffusion Policy $\pi_{dp}$ as the proposal generator, we first replace $\pi_{dp}$ with a discrete control decoder $\pi_{ctrl}'$ with a similar architecture to $\pi_{ctrl}$, which can also generate $N$ proposals based on top-$N$ confidence scores. This leads to a degradation of 5.78 DS as shown in Tab.~\ref{tab:abl_dp}, , which potentially stems from the poor ability of $\pi_{ctrl}'$ to generate long and smooth control sequences. 
Further, we investigate the ability of $\pi_{dp}$ in modeling high-frequency control sequences. Tab.~\ref{tab:abl_ctrl_dp_freq} indicates a performance degradation when using a low-frequency $\pi_{dp}$.
On the contrary, a low-frequency $\pi_{ctrl}$ is better than a high-frequency one. This echos the previous finding and proves the effectiveness of employing $\pi_{dp}$ as the generator.
Finally, Tab.~\ref{tab:abl_n_prop} examines the impact of varying the number of proposals. Increasing the proposal number $N$ from 5 to 20 leads to saturating performance on DS, while SR mildly fluctuates with a larger $N$. The fluctuation may result from over-confident matching between $\pi_{dp}$ and the other decoders, while we can potentially benefit from small differences between their predictions for exploration.

\begin{table}[t]

\centering
\small
\resizebox{\columnwidth}{!}{

\begin{tabular}{ccccc|c}
\toprule
\multicolumn{5}{c}{\textbf{Method}} & 
\textbf{Latency (ms)} \\
\midrule 
\multicolumn{5}{c|}{UniAD-Base~\cite{hu2023planning}} & 555.6  \\
\multicolumn{5}{c|}{VAD~\cite{jiang2023vad}} & 224.3   \\
\multicolumn{5}{c|}{DriveAdapter~\cite{jia2023driveadapter}} & 894.0   \\

\midrule
\textbf{Method} & \textbf{$\pi_{traj}$}& \textbf{$\pi_{ctrl}$}& \textbf{$\pi_{dp}$} & $\mathcal{F}$ & \textbf{Latency (ms)} \\
\midrule
\multirow{4}{*}{Hydra-NeXt} & \checkmark & & & & 245.4 \\
 & \checkmark & \checkmark & & & 250.6 \\
& \checkmark & \checkmark & \checkmark  & & 998.6 \\
& \checkmark & \checkmark & \checkmark & \checkmark & 528.3 \\

\bottomrule
\end{tabular}

}
\vspace{-0.1in}
\caption{
\textbf{Analysis of Runtime Efficiency}. The latency of Hydra-NeXt and VAD are benchmarked on an NVIDIA RTX 3090, while UniAD and DriveAdapter are on NVIDIA Tesla A100 and A6000, respectively.
$\mathcal{F}$ refers to Flash-attention~\cite{dao2022flashattention} for acceleration.
}
\vspace{-0.2in}
\label{tab:latency}
\end{table}
\vspace{-0.1in}
\subsection{Runtime Efficiency and Visualization}

\noindent\textbf{Runtime Efficiency}. Tab.~\ref{tab:latency} compares the runtime efficiency of \hydra with E2E baselines.
Since $\pi_{traj}$ performs planning based on perspective image tokens without constructing an explicit BEV feature~\cite{hu2023planning}, it achieves higher efficiency than the previous UniAD.
The control decoder $\pi_{ctrl}$ is also lightweight.
Nevertheless, involving the Diffusion Policy $\pi_{dp}$ for refining trajectories results in an efficiency degradation, which is an inherent limitation of the iterative denoising process~\cite{chi2023diffusion} and deserves further optimization.

\noindent\textbf{Visualization}. Fig.~\ref{fig:vis} depicts the process of Trajectory Refinement in a highway merging scenario and an unprotected left turn. Multiple control sequences are proposed by $\pi_{dp}$ and transformed into trajectories. The best-matching proposal can smooth the trajectory produced by $\pi_{traj}$ with a large curvature, ensuring kinematic feasibility.

\vspace{-0.1in}

\section{Conclusion}
\vspace{-0.05in}
In this paper, we first analyzed the development of current end-to-end autonomous driving systems, focusing on the discrepancy between open-loop and closed-loop environments and the challenges of bringing an open-loop trained model to closed-loop driving.
These challenges include handling reactive agents and adhering to kinematic constraints.
To address these issues, we propose Hydra-NeXt, a unified approach for trajectory and control prediction.
By integrating Hydra-MDP with the Multi-head Motion Decoder and the Trajectory Refinement network, Hydra-NeXt obtains stronger closed-loop driving performance.
Specifically, \hydra demonstrates superior performance of 65.89 Driving Score (DS) and 48.20\% Success Rate (SR) on the closed-loop driving benchmark Bench2Drive and outperforms previous state-of-the-art methods by a significant margin. 
\hydra also achieves a new state-of-the-art on the real-world E2E planning benchmark NAVSIM.

\noindent \textbf{Acknowledgement} This work was supported in part by the National Natural Science Foundation of China (Grant 62472098).  This work was supported by the Science and Technology Commission of Shanghai Municipality (No. 24511103100).

{
    \small
    \bibliographystyle{ieeenat_fullname}
    \bibliography{main}
}

\clearpage
\maketitlesupplementary
\begin{table*}[]
\centering
\small
\selectfont
\begin{tabular}{l|cc|ccccc|c}
\toprule
\textbf{Method} & \textbf{Perception Network} & \textbf{Grid Search} & \textbf{NC $\uparrow$} & \textbf{DAC $\uparrow$} & \textbf{TTC $\uparrow$} & \textbf{EP $\uparrow$} & \textbf{C $\uparrow$} & \textbf{PDMS $\uparrow$} \\ \midrule
Transfuser~\cite{chitta2022transfuser} & Transfuser~\cite{chitta2022transfuser} & - & 97.7& 92.8& 92.8& 79.2& \textbf{100}& 84.0 \\ 

DRAMA~\cite{yuan2024drama} & Transfuser~\cite{chitta2022transfuser}* & - & 98.0& 93.1& 94.8& 80.1& \textbf{100}& 85.5 \\ 
Hydra-MDP~\cite{li2024hydra} & Transfuser~\cite{chitta2022transfuser} & \xmark  & 97.9 & 91.7 & 92.9 & 77.6 & \textbf{100} & 83.0 \\ 
Hydra-MDP~\cite{li2024hydra} & Transfuser~\cite{chitta2022transfuser} & \cmark  & 98.3& 96.0& 94.6& 78.7& \textbf{100}& 86.5 \\ 
DiffusionDrive~\cite{liao2024diffusiondrive} & Transfuser~\cite{chitta2022transfuser} & -  & 98.2& 96.2& 94.7& \textbf{82.2}& \textbf{100}& 88.1 \\ 
\midrule
\rowcolor{gray!20}
\textbf{Hydra-NeXt} & Transfuser~\cite{chitta2022transfuser} & \xmark & \textbf{98.4}& 95.9& \textbf{94.8}& 80.6& \textbf{100}& 87.2 \\
\rowcolor{gray!20}
\textbf{Hydra-NeXt} & Transfuser~\cite{chitta2022transfuser} & \cmark & 98.1& \textbf{97.7}& 94.6& 81.8& \textbf{100}& \textbf{88.6} \\ 
\bottomrule
\end{tabular}
\caption{\textbf{Performance of E2E-AD Methods on NAVSIM.} 
All methods above use Transfuser~\cite{chitta2022transfuser} with ResNet34~\cite{he2016deep} as the perception backbone.
*DRAMA~\cite{yuan2024drama} uses Mamba~\cite{dao2024transformers} for multi-modal interaction.
Hydra-MDP~\cite{li2024hydra} uses grid search to obtain the optimal hyper-parameters for weighting different predicted metric scores.}
\label{tab:navsim_supp}
\end{table*}

\section{Details on Training Hydra-NeXt}
In this section, we provide the details on training Hydra-NeXt such as loss functions and the formulation of the diffusion policy $\pi_{dp}$. 
The three policies used in Hydra-NeXt ($\pi_{traj}$, $\pi_{ctrl}$, and $\pi_{dp}$) are all trained end-to-end simultaneously on the Bench2Drive dataset~\cite{jia2024bench}, which is collected by the RL-based expert Think2Drive~\cite{li2024think}. 
External data from different experts such as PDMLite~\cite{Beißwenger2024PdmLite} are not used.

\subsection{$\pi_{traj}$: Trajectory Decoder}
The loss of $\pi_{traj}$ is consistent with Hydra-MDP~\cite{li2024hydra}, which consists of two terms: an imitation loss $\mathcal{L}_{im}$ and a knowledge distillation loss $\mathcal{L}_{kd}$.
We denote $k$ trajectory anchors used in $\pi_{traj}$ as $\{T_i\}_{i=1}^k$.
The imitation loss is calculated as a cross entropy based on the expert trajectory $\hat{T}$ and the predicted imitation scores for each trajectory anchor $\{\mathcal{S}_i^{im}\}_{i=1}^k$:
\begin{equation}
\begin{array}{c}
    y_i=\frac{e^{-(\hat{T}-T_i)^2}}{\sum_{j=1}^{k} e^{-(\hat{T}-T_j)^2}} \\[0.05cm]
    \mathcal{L}_{im}=-\sum_{i=1}^{k} y_i\log(\mathcal{S}^{im}_i).
\end{array}
\vspace{-0.15cm}
\end{equation}
$y_i$ measures the similarity between the expert trajectory $\hat{T}$ and each trajectory anchor $T_i$.
The knowledge distillation loss aims to learn open-loop objectives via binary cross-entropy between predicted metric scores $\{\mathcal{S}_i^{m}\}_{i=1}^k$ and the ground-truth metric scores $\{\hat{\mathcal{S}}_i^{m}\}_{i=1}^k$:
\begin{equation}
\begin{aligned}
\scalebox{0.9}{
$\mathcal{L}_{kd}=-\sum_{m, i} \hat{\mathcal{S}}^m_i \log \mathcal{S}^m_i + (1-\hat{\mathcal{S}}^m_i) \log (1 - \mathcal{S}^m_i)$} \\
\text{where } m \in \{COL,SLK,EP\}.
\end{aligned}
\vspace{-0.15cm}
\end{equation}
$COL,SLK,EP$ correspond to the Collision, Soft Lane Keeping, and Ego Progress metrics defined in Sec. 3.
The ground-truth metric scores are binary values, derived from the privileged information and each trajectory anchor.
The overall loss of $\pi_{traj}$ is formulated as:
\begin{equation}
    \mathcal{L}_{traj}=\mathcal{L}_{im} + \mathcal{L}_{kd}.
\vspace{-0.15cm}
\end{equation}

\subsection{$\pi_{ctrl}$: Control Decoder}
The loss of $\pi_{ctrl}$ computes a classification-based loss for each control signal (\ie brake, throttle, and steer) across $t_{ctrl}$ timesteps. For simplicity, a control signal tuple $C=(brake, throttle, steer)$ is abbreviated as $(b, th, s)$, while the expert demonstration is denoted as $(\hat{b}, \hat{th}, \hat{s})$.
Specifically, these loss functions are formulated as: 
\begin{equation}
\begin{array}{c}
    \mathcal{L}_{brake}=\sum_{t=1}^{t_{ctrl}} Focal(b^t,\hat{b^t}) \\[0.05cm]
    \mathcal{L}_{throttle}=\sum_{t=1}^{t_{ctrl}} CE(th^t,\hat{th^t}) \\[0.05cm]
    \mathcal{L}_{steer}=\sum_{t=1}^{t_{ctrl}} CE(s^t,\hat{s^t}) \\[0.05cm]
    \mathcal{L}_{ctrl} = \mathcal{L}_{brake}+\mathcal{L}_{throttle}+\mathcal{L}_{steer},
\end{array}
\vspace{-0.15cm}
\end{equation}
where $Focal$ stands for the focal loss~\cite{lin2017focal} and $CE$ is the cross entropy loss. We empirically find that applying a focal loss to throttle and steer predictions leads to a performance degradation, possibly due to their more balanced distribution in the dataset.

\subsection{$\pi_{dp}$: Diffusion Policy}
$\pi_{dp}$ is based on the standard Diffusion Policy~\cite{chi2023diffusion} trained on continuous data.
Given expert control signals $(\hat{C^1},...,\hat{C^{t_{dp}}})$ across $t_{dp}$ frames, 
$\pi_{dp}$ is trained to predict the noise $\varepsilon^{j}$ added to the expert control signals, where $j$ is the denoising iteration.
MSE loss is applied for noise prediction: 
\begin{equation}
    \mathcal{L}_{dp} = MSE(\varepsilon^j,\pi_{dp}((\hat{C^1},...,\hat{C^{t_{dp}}})+\varepsilon^j,j)).
\end{equation}
Finally, the overall loss of Hydra-NeXt becomes
\begin{equation}
    \mathcal{L} = \mathcal{L}_{traj} + \mathcal{L}_{ctrl} + \mathcal{L}_{dp}.
\end{equation}

\section{Performance of Individual Policies}
We conduct experiments on how each individual policy performs on the Bench2Drive Benchmark. As shown in Tab.~\ref{tab:policy_supp}, using $\pi_{ctrl}$ alone leads to serious performance degradation compared with the full version of Hydra-NeXt (-16.56 DS and -29.33 SR).
Moreover, $\pi_{dp}$ achieves better results when its control candidate follows the predicted trajectory from $\pi_{traj}$ rather than being randomly selected (+5.36 DS and +4.28 SR), highlighting the importance of trajectory guidance.
Finally, the full version of Hydra-NeXt achieves substantial improvements (+13.09 DS and +17.47 SR) compared with the baseline Hydra-MDP ($\pi_{traj}$).

\begin{table}[H]
\small
\centering
\begin{tabular}{ccc|cc}
\toprule
\textbf{$\pi_{traj}$}   & \textbf{$\pi_{ctrl}$} & \textbf{$\pi_{dp}$} & \textbf{Driving Score $\uparrow$} & \textbf{Success Rate $\uparrow$} \\ \midrule
\checkmark   & & & 52.80 & 30.73 \\ 
& \checkmark & & 49.33 & 18.87  \\ 
& & $Rand.$ & 51.18 & 27.98  \\ 
& & $Traj.$ & 56.54 & 32.26  \\ 
\checkmark   & \checkmark & \checkmark & \textbf{65.89} & \textbf{48.20}  \\ 
\bottomrule
\end{tabular}
\caption{\textbf{Performance of Individual Policies on Bench2Drive.} 
}
\label{tab:policy_supp}
\end{table}

\section{Efficient Diffusion Policy $\pi_{dp}$}

\begin{table}[h]
\small
\selectfont
\centering
\begin{tabular}{l|ccc}
\toprule
\textbf{Method}  & \textbf{Latency (ms) $\downarrow$}   & \textbf{D.S. $\uparrow$} & \textbf{S.R. $\uparrow$} \\ \midrule
VAD & 224.3 & 39.42 & 10.00 \\ 
Hydra-NeXt (w/o $\pi_{dp}$) & 250.6 & 60.40 & 48.10 \\ 
Hydra-NeXt* (DDPM)   & 528.3 & 65.89 & 48.20 \\ 
Hydra-NeXt* (DDIM)   & 243.9 & 64.87 & 46.63 \\ 
\bottomrule
\end{tabular}
\caption{\textbf{Efficiency of Different Diffusion Schedulers.} * denotes Flash-attention~\cite{dao2022flashattention}.}
\end{table}
We found that the efficiency of $\pi_{dp}$ can be greatly enhanced without sacrificing too much performance. By replacing the DDPM with a DDIM scheduler~\cite{song2020denoising}, we can reduce the latency by approximately 53\%, decreasing the number of denoising steps from 100 to 20. Additionally, by combining this modification with flash attention, we achieve latencies comparable to VAD while maintaining strong closed-loop performance, with only a marginal 1\% drop in the Driving Score. Therefore, we conclude that $\pi_{dp}$ can be both effective and efficient with the right design choices.

\section{Implementation on NAVSIM}
Our implementation of Hydra-NeXt on NAVSIM uses a different perception network Transfuser~\cite{chitta2022transfuser} following Hydra-MDP~\cite{li2024hydra}. 
Transfuser~\cite{chitta2022transfuser} features two backbones for camera and lidar feature extraction, a BEV segmentation head, a 3D object detection head, and transformer layers for multi-modal feature interaction.
This setting helps to make a fair comparison to baselines such as Hydra-MDP and DiffusionDrive~\cite{liao2024diffusiondrive}.
For $\pi_{ctrl}$ and $\pi_{dp}$, we incorporate the same transformer architectures used on Bench2Drive for acceleration and steering rate predictions since NAVSIM~\cite{Dauner2024NEURIPS} does not utilize control signals like CARLA~\cite{CARLAv2, dosovitskiy2017carla} and only evaluates the trajectory.
Therefore, these auxiliary predictions only act as extra learning targets.
As a result, Hydra-NeXt surpasses the state-of-the-art planner DiffusionDrive by 0.5 PDMS (see Tab.~\ref{tab:navsim_supp}) when adopting the grid search trick~\cite{li2024hydra} among different metric scores in $\pi_{traj}$.

\section{Implementation on CARLA-Garage}

\begin{table}[ht]
\small
\selectfont
\centering
\begin{tabular}{c|c|cc}
\toprule
\textbf{Method} & \textbf{Expert} & \textbf{D.S. $\uparrow$} & \textbf{S.R. $\uparrow$} \\ \midrule
Hydra-NeXt & Think2Drive & 73.86 & 50.00  \\ 
TF++\dag~\cite{Jaeger2023ICCV} & PDM-Lite & 84.21 & 67.27 \\ 
SimLingo~\cite{renz2025simlingo} & PDM-Lite & 85.94 & 66.82  \\ 
Hydra-NeXt*\dag & PDM-Lite & 86.00 & 68.18  \\ 
\bottomrule
\end{tabular}
\caption{\textbf{Performance of Models Trained with Different Experts.} \dag  Ensemble of three models trained with different seeds.}

\end{table}

We provide additional results on the CARLA-Garage dataset~\cite{Jaeger2023ICCV} collected by the PDM-Lite expert~\cite{Beißwenger2024PdmLite}.
Compared to the Bench2Drive dataset~\cite{jia2024bench}, this dataset features smooth driving with no jittery behavior.
We experiment with $\pi_{traj}$ and $\pi_{dp}$ within the TF++ framework~\cite{Jaeger2023ICCV}.
Specifically, we replace the longitudinal control head of TF++ with $\pi_{traj}$ (imitation + collision head) and fuse the outputs of $\pi_{dp}$ with longitudinal and lateral controls.
We then ensemble three models trained with different seeds. This variant, Hydra-NeXt*, achieves a Driving Score of 86\%, surpassing existing methods.

\section{Visualization Results}
\begin{figure*}[tp]
    \centering
        \includegraphics[width=0.95\linewidth]{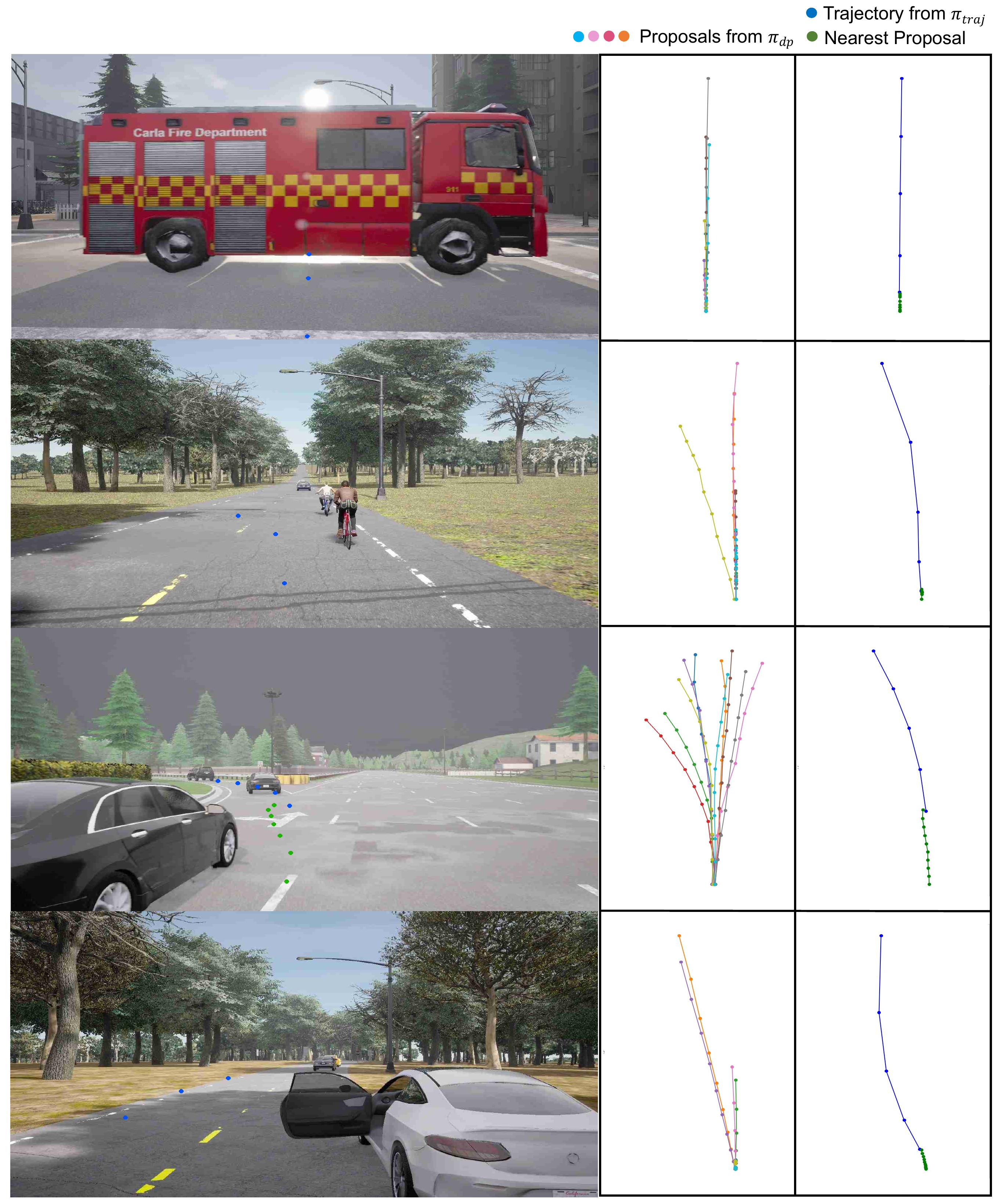}
    \caption{\textbf{More Visualizations of Trajectory Refinement.} The figure shows the front-view image, trajectories transformed from control proposals, and the selected nearest proposal to the predicted trajectory.
    }
    \label{fig:vis_supp}
\end{figure*}

Fig.~\ref{fig:vis_supp} shows more visualization results in interactive scenarios (Merging, Overtaking, and Give Way).
The Diffusion Policy $\pi_{dp}$ can capture multiple planning modes such as following other agents or overtaking them (see the second row and the fourth row of the figure).

\section{Limitations.} 
Although Hydra-NeXt shows outstanding closed-loop driving performance compared with E2E methods, it still falls behind RL-based experts using privileged input. The runtime efficiency of the Diffusion Policy also deserves optimization. We expect these to be addressed in future research.

\end{document}